\documentclass[journal]{IEEEtran}
\usepackage{graphicx}
\usepackage{stfloats}
\graphicspath{{fig/}}

\usepackage{microtype}
\usepackage[cmex10]{amsmath}
\usepackage{amssymb}
\usepackage{cancel}
\DeclareMathOperator*{\argmax}{arg\,max}
\newcommand{\T}{^\textrm{T}}

\renewcommand{\vec}[1]{\boldsymbol{\mathbf{#1}}}
\newcommand{\defeq}{\triangleq}

\usepackage{tabularx}
\usepackage{multirow}
\newcommand{\mytable}{
    \centering
    \renewcommand{\arraystretch}{1.2}
    }
\newcolumntype{C}{>{\centering\arraybackslash}X}
\newcolumntype{L}{>{\raggedright\arraybackslash}X}

\usepackage{algorithm}  % should go before \usepackage{hyperref}

\usepackage{url}

\usepackage{algpseudocode}  % should go after \usepackage{hyperref}

\usepackage{cite}  % automatically reorder inline citations

\usepackage{color}

\begin{document}

\title{Unsupervised Word Segmentation and Lexicon Discovery Using Acoustic Word Embeddings}
\author{Herman~Kamper,~\IEEEmembership{Student Member,~IEEE,}
        Aren~Jansen,~\IEEEmembership{Member,~IEEE,}
        and~Sharon Goldwater%,~\IEEEmembership{}% <-this % stops a space
\thanks{H. Kamper is with the School of Informatics, University of Edinburgh, UK  (email: see http://www.kamperh.com).}%
\thanks{A. Jansen performed this work while with the Human Language Technology Center of Excellence at Johns Hopkins University, USA (email: aren@jhu.edu).}%
\thanks{S. J. Goldwater is with the School of Informatics, University of Edinburgh, UK (email: sgwater@inf.ed.ac.uk).}%
\thanks{HK is funded by a Commonwealth Scholarship.  This work was supported in part by a James S.\ McDonnell Foundation Scholar Award to SG. }
}

% Paper headers
\markboth{Accepted to the IEEE Transactions on Audio, Speech, and Language Processing, 2016}
{Kamper, Jansen and Goldwater}
% \IEEEpubid{0000--0000/00\$00.00~\copyright~2014 IEEE}
\IEEEpubid{\copyright~2016 IEEE}

\maketitle

\begin{abstract}
In settings where only unlabelled speech data is available, speech technology needs to be developed without transcriptions, pronunciation dictionaries, or language modelling text.  A similar problem is faced when modelling infant language acquisition.  In these cases, categorical linguistic structure needs to be discovered directly from speech audio.  We present a novel unsupervised Bayesian model that segments unlabelled speech and clusters the segments into hypothesized word groupings.  The result is a complete unsupervised tokenization of the input speech in terms of discovered word types.  In our approach, a potential word segment (of arbitrary length) is embedded in a fixed-dimensional acoustic vector space.  The model, implemented as a Gibbs sampler, then builds a whole-word acoustic model in this space while jointly performing segmentation.
We report word error rates in a small-vocabulary connected digit recognition task by mapping the unsupervised decoded output to ground truth transcriptions.
The model achieves around 20\% error rate, outperforming a previous HMM-based system by about 10\% absolute.  Moreover, in contrast to the baseline, our model does not require a pre-specified vocabulary size.
\end{abstract}


\IEEEpeerreviewmaketitle

\section{Introduction}

\IEEEPARstart{G}{reat} advances have been made in speech recognition in the last few years.
However, most of these improvements have come from supervised techniques, relying on large corpora of transcribed speech audio data, texts for language modelling, and pronunciation dictionaries.
For under-resourced languages, only limited amounts of these resources are available.
In the extreme \textit{zero-resource} case, only raw speech audio is available for system development.
In this setting, unsupervised methods are required to discover linguistic structure directly from audio.
Similar techniques are also necessary to model how infants acquire language from speech input in their native language.

Researchers in the speech processing community have recently started to use completely unsupervised techniques to build zero-resource technology directly from unlabelled speech data.
Examples include the query-by-example systems of~\cite{zhang+glass_asru09,zhang+etal_icassp12,metze+etal_icassp13,levin+etal_icassp15}, and the unsupervised term discovery (UTD) systems of~\cite{park+glass_taslp08,jansen+vandurme_asru11}, which aim to find repeated words or phrases in a speech collection.
Few studies, however, have considered an unsupervised system able to perform a full-coverage segmentation of speech into
word-like units---the goal of this paper.
Such a system would perform {fully unsupervised speech recognition}, allowing downstream applications, such as query-by-example search and speech indexing (grouping together related utterances in a corpus), to be developed in a manner similar to when supervised systems are available.

Another community that would have significant interest in such a system is the scientific cognitive modelling community.
Here, researchers are interested in the problems faced during early language learning: infants have to learn phonetic categories and a lexicon for
their native language using speech audio as input~\cite{rasanen_speechcom12}.
In this community, unsupervised models have been developed that perform
full-coverage word segmentation of data into a sequence of words, proposing word boundaries
for the entire input.
However, these models take 
transcribed symbol sequences as input, rather than continuous speech~\cite{goldwater+etal_cognition09}.

A few recent studies~\cite{sun+vanhamme_csl13,chung+etal_icassp13,walter+etal_asru13,lee_phd14}, summarized in detail in Section~\ref{sec:related_studies_full_coverage}, share our goal of full-coverage speech segmentation. 
Most of these
follow an approach of phone-like subword discovery with subsequent or joint word discovery, working directly on the frame-wise acoustic speech features.

The model we present is a novel Bayesian model that jointly segments speech data into word-like \textit{segments} and then clusters these segments, each cluster representing a discovered word type.\footnote{`Word type' refers to distinct words, i.e.\ the entries in a lexicon, while `word token' refers to different realizations of a particular word.}
Instead of operating directly on acoustic frames, our model uses a fixed-dimensional representation of whole segments: any potential word segment of arbitrary length is mapped to a fixed-length vector, its \textit{acoustic embedding}.
Because the model has no subword level of
representation and models whole segments directly, we refer to the
model as \textit{segmental}.\footnote{{
`Segmental' is used here, as in~\cite{zweig+nguyen_interspeech10}, to distinguish approaches
operating on whole units of speech from those doing frame-wise modelling. This is
different from the traditional linguistic usage of `{segment}' to refer to phone-sized units.
}}
Using these fixed-dimensional acoustic
embeddings,
we extend the Bayesian segmentation model of Goldwater et al.~\cite{goldwater+etal_cognition09} (which took symbolic input) to the continuous speech domain.
In an evaluation on an unsupervised digit recognition task using the TIDigits corpus, 
we show that our model outperforms the unsupervised HMM-based model of Walter et al.~\cite{walter+etal_asru13}, without specifying the vocabulary size and without relying on a UTD system for model initialization.

\IEEEpubidadjcol
The main contribution of this work is to introduce a novel segmental Bayesian model for unsupervised segmentation and clustering of speech into hypothesized words---an approach which is distinct from any presented before.
Our preliminary work in this direction was presented in~\cite{kamper+etal_interspeech15}.
Here we present a complete mathematical description of the model
and much more extensive experiments and discussion. In particular, we
provide a thorough analysis of
the discovered structures and model errors, investigate the effects of model hyperparameters,
and discuss the challenges involved in scaling our
approach to larger-vocabulary tasks.

\section{Related Work}

In the following we describe relevant studies from both the speech processing and cognitive modelling communities.

\subsection{Discovery of words in speech}
\label{sec:review_utd}

Unsupervised term discovery (UTD), sometimes referred to as `lexical
discovery' or `spoken term discovery', is the task of finding
meaningful repeated word- or phrase-like patterns in raw speech audio.
Most state-of-the-art UTD systems are based on the seminal work of
Park and Glass~\cite{park+glass_taslp08}, who proposed a method to find pairs of similar
audio segments and then cluster them into hypothesized word types.
The pattern matching step uses a variant of dynamic time warping (DTW)
called segmental DTW, which allows similar sub-sequences within two
vector time series to be identified, rather than comparing entire
sequences as in standard DTW. Follow-up work has built on Park and
Glass' original method in various ways, for example through improved
feature representations~\cite{zhang+glass_asru09,zhang+etal_icassp12} or by greatly improving its efficiency~\cite{jansen+vandurme_asru11}.

Like our own system, many of these UTD systems operate on whole-word
representations, with no subword level of representation. However,
each word is represented as a vector time series with variable
dimensionality (number of frames), requiring DTW for comparisons.
Since our own system uses fixed-dimensional word representations, we can define an acoustic model over these embeddings and make comparisons without requiring any alignment.
In addition, UTD systems aim to find and
cluster repeated, isolated acoustic segments, leaving much of the
input data as background. In contrast, we aim for full-coverage segmentation of the entire speech input into
hypothesized~words.

\subsection{Word segmentation of symbolic input}

Cognitive scientists have long been interested in how infants
learn to segment words and discover the lexicon of their native
language, with computational models seen as one way to specify and
test particular theories (see \cite{rasanen_speechcom12,goldwater+etal_cognition09}  for reviews).
In this community, most computational models of word segmentation
perform full-coverage segmentation of the data into a sequence of
words.
However,
these models generally take phonemic or phonetic strings as input,
rather than continuous~speech.

Early word segmentation approaches using phonemic input include those based
on transition probabilities~\cite{brent_ml99}, neural networks~\cite{christiansen+etal_lcp98} and
probabilistic models~\cite{venkataraman_cl01}. The model presented here is based on the
non-parametric Bayesian approach of Goldwater et al.~\cite{goldwater+etal_cognition09}, which was
shown to yield more accurate segmentations than previous work.
Their approach learns a language model over the tokens in
its inferred segmentation, incorporating priors that favour predictable
word sequences and a small vocabulary.\footnote{
They experimented with learning either a unigram or bigram language model, and found that the proposed boundaries of both models were very accurate, but the unigram model proposed too few boundaries.
}
The original method uses a Gibbs sampler to sample individual boundary
positions; our own sampler is based on the later work of Mochihashi et
al.~\cite{mochihashi+etal_acl09} who presented a blocked sampler that uses dynamic programming
to resample the segmentation of a full utterance at once.

Goldwater et al.'s original model assumed that every instance of a
word is represented by the same sequence of phonemes; later
studies~\cite{neubig+etal_interspeech10,elsner+etal_emnlp13,heymann+etal_asru13} proposed noisy-channel extensions in order to deal
with variation in word pronunciation.  Our
model can also be viewed as a noisy-channel extension to the original
model, but with a different type of channel model. In~\cite{neubig+etal_interspeech10,elsner+etal_emnlp13,heymann+etal_asru13},
variability is modeled symbolically as the conditional probability of
an output phone given the true phoneme (so the input to the models is
a sequence or lattice of phones), whereas our channel model is a true acoustic
model (the input is the speech signal).  As in the
phonetic noisy channel model of~\cite{elsner+etal_emnlp13}, we learn the language model and channel
model jointly.

\subsection{Full-coverage segmentation of speech}
\label{sec:related_studies_full_coverage}

We highlight four recent studies that share our goal of full-coverage word segmentation of speech.

Sun and Van hamme~\cite{sun+vanhamme_csl13} developed an approach based on non-negative matrix factorization (NMF).
NMF is a technique which allows fixed-dimensional representations of speech utterances (typically co-occurrence statistics of acoustic events) to be factorized into lower-dimensional parts, corresponding to phones or words~\cite{stouten+etal_ieee08}. 
To capture temporal information, Sun and Van hamme~\cite{sun+vanhamme_csl13} incorporated NMF in a maximum likelihood training procedure for discrete-density HMMs.
They applied this approach to an 11-word unsupervised connected digit recognition task using the TIDigits corpus.
They learnt 30 unsupervised HMMs, each representing a discovered word type. 
They found that the discovered word clusters corresponded to sensible words or subwords:
average cluster purity 
was around 85\%.
Although NMF itself relies on a fixed-dimensional representation (as our system does) the final HMMs of their approach still perform frame-by-frame modelling (as also in  the studies below).
Our approach, in contrast, operates directly on a fixed-dimensional representation of speech segments.

Chung et al.~\cite{chung+etal_icassp13} used an HMM-based approach which alternates between subword and word discovery.
Their system models discovered subword units as continuous-density HMMs and learns a lexicon in terms of these units by alternating between unsupervised decoding and parameter re-estimation.
For evaluation, the output from their unsupervised system was compared to the ground truth transcriptions and every discovered word type was mapped to the ground truth label that resulted in the smallest error rate.
This allowed their system to be evaluated in terms of unsupervised WER; on a four-hour Mandarin corpus with a vocabulary size of about 400, they achieved WERs around 60\%.

Lee et al.~\cite[Ch.~3]{lee_phd14},\cite{lee+etal_tacl15} developed a non-parametric hierarchical Bayesian model for full-coverage speech segmentation.
Using adaptor grammars (a generalized framework for defining such Bayesian models), an unsupervised subword acoustic model developed in earlier work~\cite{lee+glass_acl12} was extended with syllable and word layers, as well as a noisy channel model for capturing phonetic variability in word pronunciations.
When applied to speech from single speakers in the MIT Lecture corpus, most of the words with highest TF-IDF scores were successfully discovered,
and Lee et al.\ showed that joint modelling of subwords, syllables and words improved term discovery performance.
{In~\cite{lee+etal_tacl15}, although unsupervised WER was not reported, the full-coverage segmentation of the system was evaluated in terms of word boundary $F$-score.}
% Unfortunately, although the system performs full-coverage segmentation, no evaluation was provided of this aspect (only discovery of key terms was evaluated).
As in these studies, we also follow a Bayesian approach.
However, our model operates directly at the whole-word level instead of having a hierarchy of layers from words down to acoustic features. %, as is done in Lee's model.
In addition, we evaluate on a small-vocabulary multi-speaker corpus rather than large-vocabulary single-speaker data.

The work that is most directly comparable to our own is that of Walter et al.~\cite{walter+etal_asru13}.
They developed a fully unsupervised system for connected digit recognition, also using the TIDigits corpus.
As in~\cite{chung+etal_icassp13}, they followed a two-step iterative approach of subword and word discovery.
For subword discovery, speech is partitioned into subword-length segments and clustered based on DTW similarity.
For every subword cluster, a continuous-density HMM is trained.
Word discovery takes as input the subword tokenization of the input speech.
Every word type is modelled as a discrete-density HMM with multinomial emission distributions over subword units, accounting for noise
and pronunciation variation.
HMMs are updated in an iterative procedure of parameter estimation and decoding.
Eleven of the whole-word HMMs were trained, one for each of the digits in the corpus.
Using a random initialization, their system achieved an unsupervised WER of 32.1\%; using UTD~\cite{park+glass_taslp08} to provide initial word identities and boundaries, 18.1\% was achieved.
In a final improvement, the decoded output was used to train from scratch standard continuous-density whole-word HMMs. 
This led to further improvements by leveraging the well-developed HMM tools used for supervised speech recognition.

This study of Walter et al.\ shows that unsupervised multi-speaker speech
recognition on a small-vocabulary task is possible. It also provides
useful baselines on a standard dataset, and gives a reproducible evaluation method in terms of the standard WER.
Our model is comparable to Walter et al.'s word discovery system before the refinement using a traditional HMM-GMM recognizer.
We therefore use the results they obtained before refinement as baselines in our experiments.
It would be possible to apply the same refinement step to our model, but we have not done so here.

{
Most of the above studies perform explicit subword modelling, while our approach operates on fixed-dimensional embeddings of whole-word segments.
We do not argue  that the latter is necessarily superior, but rather see our approach as a new contribution;
direct whole-word modelling has both advantages and disadvantages.
On the positive side, it is often easier to identify cross-speaker similarities between words than between subwords~\cite{jansen+etal_icassp13b}, which is why most UTD systems focus on longer-spanning patterns.
And from a cognitive perspective, there is evidence that infants are able to segment whole words from continuous speech before phonetic contrasts in their native language have been fully learned~\cite{bortfeld+etal_psychol05,feldman+etal_ccss09}.
On the other hand, direct whole-word modelling in our approach makes it more difficult to explicitly include intermediate modelling layers (phones, syllables, morphemes) as Lee et al.\ did.
Furthermore, our whole-word approach is completely reliant on the
quality of the embeddings; in Section~\ref{sec:scaling} we show that
the embedding function we use deals poorly with short segments.
Improved embedding techniques are the subject of current research~\cite{kamper+etal_arxiv15} and it would be straightforward to replace the current embedding
approach with any other (including one that incorporates subword
modelling).
}

\begin{figure*}[!t]
    \centering
    {\footnotesize(a)}~\includegraphics[scale=0.725]{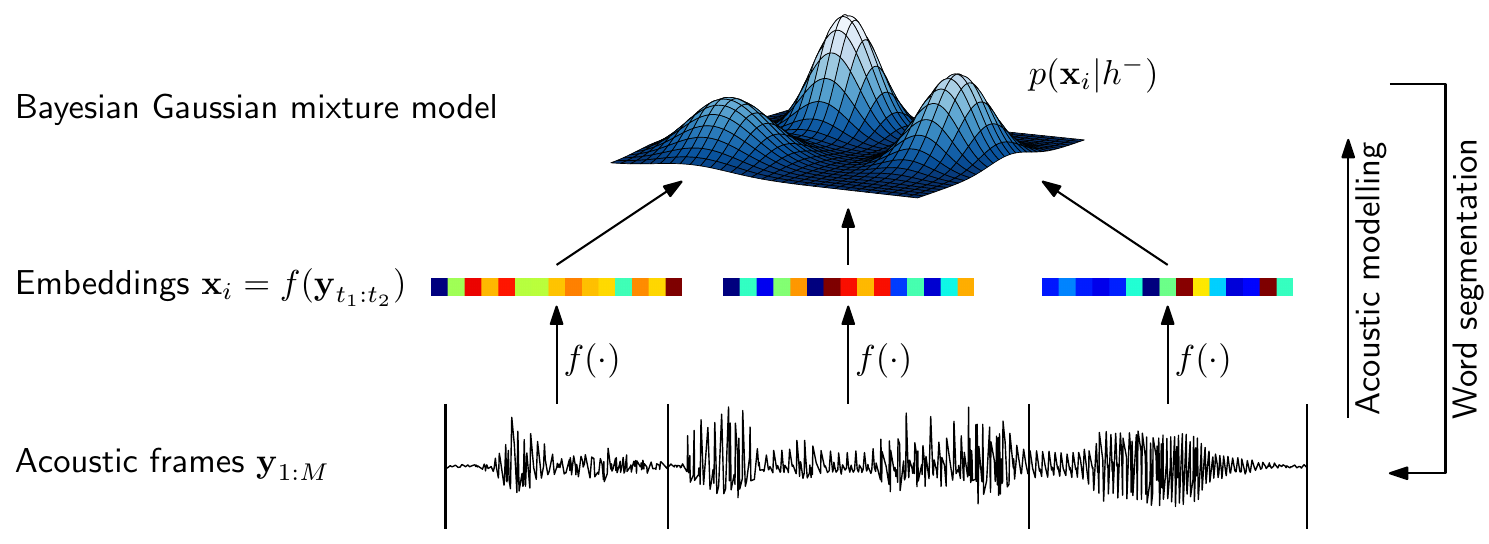} \hspace{2em} {\footnotesize(b)}~\includegraphics[scale=0.725]{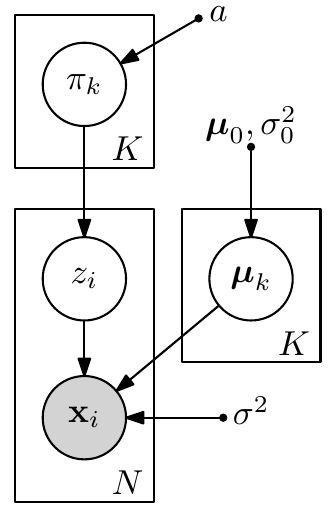}
    \caption{(a) Overview of the segmental Bayesian model for unsupervised segmentation and clustering of speech.
    (b) The graphical model of the Bayesian Gaussian mixture model with fixed spherical covariance used as acoustic model.
    }
    \label{fig:unsup_wordseg}
\end{figure*}

\section{The Segmental Bayesian Model}
\label{sec:segmental_bayesian_model}

In our approach, any potential word segment (of arbitrary length) is mapped to a vector in a fixed-dimensional space $\mathbb{R}^D$.
The goal of this \textit{acoustic word embedding} procedure is that word instances of the same type should lie close together in this space.
The different hypothesized word types are then modelled in this $D$-dimensional space using a Gaussian mixture model (GMM) with Bayesian priors.
Every mixture component of the GMM corresponds to a discovered type; the component mean can be seen as an average embedding for that word.
However, since the model is unsupervised, we do not know the identities of the true word types to which the components correspond.

Assume for the moment such an ideal GMM exists.
This Bayesian GMM is the core component in our overall approach, which is illustrated in Fig.~\ref{fig:unsup_wordseg}(a).
Given a new unsegmented unlabelled utterance of acoustic feature frames $\vec{y}_{1:M} = \vec{y}_1, \vec{y}_2, \ldots, \vec{y}_M$, the aim is to hypothesize where words start and end in the stream of features, and to which word type (GMM mixture component) every word segment belongs.
Given a proposed segmentation hypothesis (Fig.~\ref{fig:unsup_wordseg}(a) bottom), we can calculate the acoustic embedding vector for every proposed word segment (Fig.~\ref{fig:unsup_wordseg}(a) middle), calculate a likelihood score for each embedding under the current GMM (Fig.~\ref{fig:unsup_wordseg}(a) top), and obtain an overall
score for the current segmentation hypothesis.
The aim then is to find the optimal segmentation under the current GMM, which can be done using dynamic programming.
In our model, we sample a likely segmentation with a dynamic programming Gibbs sampling algorithm using
the probabilities we obtain from the Bayesian GMM. The result is a complete segmentation of the input
utterance and a prediction of the component to which every word segment belongs. %Using

In our actual model, the Bayesian GMM is built up jointly while performing segmentation: the GMM provides the likelihood terms required for segmentation, while the segmentation hypothesizes the boundaries for the word segments which are then clustered using the GMM.
The GMM (details in Section~\ref{sec:gmm}) can thus be seen as an acoustic model which discovers the underlying word types of a language, while the segmentation component (Section~\ref{sec:word_segmentation}) discovers where words start and end.
Below we provide complete details of the model.

\subsection{Fixed-dimensional representation of speech segments}
\label{sec:embeddings}

Our model requires that any acoustic speech segment in an utterance be embedded in a fixed-dimensional space.
In principle, any approach that is able to map an arbitrary-length vector time series to a fixed-dimensional vector can be used.
Based on previous results, we follow the embedding approach developed by Levin et al.~\cite{levin+etal_asru13}, as summarized below.

The notation $Y = \vec{y}_{1:T}$ is used to denote a vector time series, where each $\vec{y}_t$ is the frame-level acoustic features (e.g.\ MFCCs).
We need a mapping function $f(Y)$ that maps time series $Y$ into a space $\mathbb{R}^D$ in which proximity between mappings indicates similar linguistic content, so 
embeddings of word tokens of the same type will be close together.
In~\cite{levin+etal_asru13}, the mapping $f$ is performed as follows.
For a target speech segment, a reference vector is constructed by calculating the DTW alignment cost to every exemplar in a reference set $\mathcal{Y}_{\text{ref}} = \{ Y_i \}_{i = 1}^{N_\text{ref}}$.
Applying dimensionality reduction to the reference vector yields the embedding in $\mathbb{R}^D$.
Dimensionality reduction is performed using Laplacian eigenmaps~\cite{belkin+niyogi_neurocomp03}.

Intuitively, Laplacian eigenmaps tries to find an optimal non-linear mapping such that the $k$-nearest neighbouring speech segments in the reference set $\mathcal{Y}_{\text{ref}}$ are mapped to similar regions in the target space $\mathbb{R}^D$.
To embed an arbitrary segment $Y$ which is not an element of $\mathcal{Y}_{\text{ref}}$, a kernel-based out-of-sample extension is used~\cite{belkin+etal_jmlr06}.
This performs a type of interpolation using the exemplars in $\mathcal{Y}_{\text{ref}}$ that are similar to target segment~$Y$.

In all experiments we use a radial basis function kernel:
\begin{equation}
    K(Y_i, Y_j) = \exp \left\{ - \frac{ \left[\text{DTW} (Y_i, Y_j) \right]^2 }{2 \sigma_K^2} \right\}
\end{equation}
where $\text{DTW} (Y_i, Y_j)$ denotes the DTW alignment cost between segments $Y_i$ and $Y_j$, and $\sigma_K$ is the kernel width parameter.
In~\cite{belkin+etal_jmlr06}, it was shown that the optimal projection to the $j^{\text{th}}$ dimension in the target space is given by
\begin{equation}
    h_j(Y) = \sum_{i = 1}^{N_\text{ref}} \alpha_i^{(j)} K(Y_i, Y)
    \label{eq:embed1}
\end{equation}
The $\alpha_i^{(j)}$ terms are the solutions to the generalized eigenvector problem $(\vec{L}\vec{K} + \xi \vec{I}) \vec{\alpha} = \lambda \vec{K} \vec{\alpha}$, with $\vec{L}$ the normalized graph Laplacian, $\vec{K}$ the Gram matrix with elements $K_{ij} = K(Y_i, Y_j)$ for $Y_i, Y_j \in \mathcal{Y}_{\text{ref}}$, and $\xi$ a regularization parameter. An arbitrary speech segment $Y$ is then mapped to the embedding $\vec{x} \in \mathbb{R}^D$ given by $\vec{x} = {f} (Y) = \left[ h_1(Y), h_2(Y), \ldots, h_d(Y) \right]\T$.

We have given only a brief outline of the embedding method here; complete details can be found in~\cite{levin+etal_asru13,belkin+niyogi_neurocomp03,belkin+etal_jmlr06}.

\subsection{Acoustic modelling: discovering word types}
\label{sec:gmm}

Given a segmentation hypothesis of a corpus (indicating where words start and end), the acoustic model needs to cluster the hypothesized word segments (represented as fixed-dimensional vectors) into groups of hypothesized word types.
Note again that acoustic modelling is performed jointly with word segmentation (next section), 
but here we describe the acoustic model under the current segmentation hypothesis.
Formally, given the embedded word vectors $\mathcal{X} = \{ \vec{x}_i\}_{i = 1}^{N}$ from the current segmentation hypothesis, the acoustic model needs to assign each vector $\vec{x}_i$ to one of $K$ clusters. 

We choose for the acoustic model a 
Bayesian GMM with fixed spherical covariance.
This model treats its mixture weights and component means as random variables rather than point estimates as is done in a regular GMM.
In~\cite{kamper+etal_slt14} we showed that the Bayesian GMM performs significantly better in clustering word embeddings than a regular GMM trained with expectation-maximization.
The former also fits naturally within the sampling framework of our complete model.

The Bayesian GMM is illustrated in Fig.~\ref{fig:unsup_wordseg}(b).
For each observed embedding $\vec{x}_i$, latent variable $z_i$ indicates the component to which $\vec{x}_i$ belongs.
The prior probability that $\vec{x}_i$ belongs to component $k$ is $\pi_k = P(z_i = k)$.
Given $z_i = k$, $\vec{x}_i$ is generated by the $k^\text{th}$ Gaussian mixture component with mean vector $\vec{\mu}_k$.
All components share the same fixed covariance matrix $\sigma^2 \vec{I}$; preliminary experiments, based on~\cite{kamper+etal_slt14}, indicated that it is sufficient to only model component means while keeping covariances fixed.
Formally, the model is then defined~as:

\noindent
\begin{minipage}{.45\linewidth}
    \centering
    \begin{alignat}{2}
        &\vec{\pi}  &&\sim \textrm{Dir}\left( a/{K} \vec{1} \right) \label{eq:fbgmm1} \\
        &z_i &&\sim \vec{\pi}  \label{eq:fbgmm2}
    \end{alignat}
\end{minipage}
\hfill
\noindent
\begin{minipage}{.45\linewidth}
    \centering
    \begin{alignat}{2}
        &\vec{\mu}_k  &&\sim \mathcal{N} (\vec{\mu}_0, \sigma_0^2 \vec{I})  \label{eq:fbgmm3} \\
        &\vec{x}_i &&\sim \mathcal{N} (\vec{\mu}_{z_i}, \sigma^2 \vec{I})  \label{eq:fbgmm4}
    \end{alignat}
\end{minipage}
\vspace{\the\belowdisplayskip}

We use a symmetric Dirichlet prior in~\eqref{eq:fbgmm1} since it is conjugate to the categorical distribution in~\eqref{eq:fbgmm2}~\cite[p.~171]{barber}, and a spherical-covariance Gaussian prior in~\eqref{eq:fbgmm3} since it is conjugate to the Gaussian distribution in~\eqref{eq:fbgmm4}~\cite{murphy_bayesgauss07}.
We use $\vec{\beta} = (\vec{\mu}_0, \sigma_0^2, \sigma^2)$ to denote all the hyperparameters of the mixture components.

Given $\mathcal{X}$, we infer the component assignments $\vec{z} =
(z_1, z_2, \ldots, z_N)$ using a collapsed Gibbs sampler~\cite{resnik+hardisty_gibbs_tutorial10}.
Since we chose conjugate priors, we can marginalize over $\vec{\pi}$ and $\left\{ \vec{\mu}_k \right\}_{k = 1}^K$ and only need to sample $\vec{z}$.
This is done in turn for each $z_i$ conditioned on all the other current component assignments:
\begin{align}
    &P(z_i = k|\vec{z}_{\backslash i}, \mathcal{X} ; a, \vec{\beta} ) \nonumber \\
    &\qquad\propto P(z_i = k|\vec{z}_{\backslash i}; a)  p(\vec{x}_i|\mathcal{X}_{\backslash i}, z_i = k, \vec{z}_{\backslash i}; \vec{\beta})
    \label{eq:collapsed1}
\end{align}
where $\vec{z}_{\backslash i}$ is all latent component assignments excluding $z_i$ and $\mathcal{X}_{\backslash i}$ is all embedding vectors apart from $\vec{x}_i$.

By marginalizing over $\vec{\pi}$, the first term on the right hand side of \eqref{eq:collapsed1} can be calculated as:
\begin{equation}
    P(z_i = k|\vec{z}_{\backslash i}; {a}) = \frac{N_{k\backslash i} + a/K}{N + a - 1}
    \label{eq:first_term7}
\end{equation}
where $N_{k \backslash i}$ is the number of embedding vectors from mixture component $k$ without taking $\vec{x}_i$ into account~\cite[p.~843]{murphy}.
This term can be interpreted as a discounted unigram language modelling probability.
Similarly, it can be shown that by marginalizing over $\vec{\mu}_k$, the second term %in \eqref{eq:collapsed1}
\begin{equation}
    p(\vec{x}_i|\mathcal{X}_{\backslash i}, z_i = k, \vec{z}_{\backslash i}; \vec{\beta}) = p(\vec{x}_i | \mathcal{X}_{k \backslash i}; \vec{\beta})
    \label{eq:second_term1}
\end{equation}
is the posterior predictive of $\vec{x}_i$ for a Gaussian distribution with known spherical covariance and a conjugate prior over its means, which is itself a spherical covariance Gaussian distribution~\cite{murphy_bayesgauss07}.
Here, $\mathcal{X}_{k \backslash i}$ is the set of embedding vectors assigned to component $k$ without taking $\vec{x}_i$ into account.
Since the multivariate distributions in~\eqref{eq:fbgmm3} and~\eqref{eq:fbgmm4} have known spherical covariances, the probability density function (PDF) of the multivariate posterior predictive simply decomposes into the product of univariate PDFs; for a single dimension $x_i$ of vector $\vec{x}_i$, this PDF is given by
\begin{equation}
    p(x_i|\mathcal{X}_{k \backslash i})
    = \mathcal{N} (x_i|\mu_{N_{k \backslash i}}, \sigma_{N_{k \backslash i}}^2 + \sigma^2) \label{eq:univariate_post_predict}
\end{equation}
where
\begin{equation}
    \sigma_{N_{k \backslash i}}^2 = \frac{\sigma^2\sigma_0^2}{N_{k \backslash i}\sigma_0^2 + \sigma^2} \text{\ \ ,\ \ }
    \mu_{N_{k \backslash i}} = \sigma_{N_{k \backslash i}}^2 \left( \frac{\mu_0}{\sigma_0^2} + \frac{N_{k \backslash i}\overline{x}_{k \backslash i}}{\sigma^2} \right)
\end{equation}
and $\overline{x}_{k \backslash i}$ is component $k$'s sample mean for this dimension~\cite{murphy_bayesgauss07}.

Although we use a model with a fixed number of components $K$,
Bayesian models that marginalize over their parameters have been shown
to prefer sparser solutions than maximum-likelihood models with the
same structure\cite{goldwater+griffiths_acl07}.  Thus, our Bayesian GMM tends towards
solutions where most of the data are clustered into just a few
components, and we can find good minimally constrained solutions by
setting $K$ to be much larger than the expected true number of
types and letting the model decide how many of those components
to use.

\subsection{Joint segmentation and clustering}
\label{sec:word_segmentation}

The acoustic model of the previous section can be used to cluster
existing segments. Our joint segmentation and clustering system works
by first sampling a segmentation of the current utterance based on the
current acoustic model (marginalizing over cluster assignments for
each potential segment), and then resampling the clusters of the newly
created segments. The inference algorithm is a blocked Gibbs sampler
using dynamic programming, based on the work of Mochihashi et al.~\cite{mochihashi+etal_acl09}.

\begin{figure}[!b]
\hrulefill
\begin{algorithmic}[1]
\small
\State Choose an initial segmentation (e.g.\ random).
\For{$j = 1$ to $J$}\Comment{Gibbs sampling iterations}

    \For{$i = $ randperm$(1$ to $S)$} \Comment{Select utterance $\vec{s}_i$}
    
        \State Remove embeddings $\mathcal{X}(\vec{s}_i)$ from acoustic model.\label{alg_line:remove_embeds}
        
        \State Calculate $\alpha$'s using~\eqref{eq:forward}.
        \label{alg_line:forward}
        
        \State Draw $\mathcal{X}(\vec{s}_i)$ by sampling word boundaries using~\eqref{eq:backward}. \label{alg_line:sample_bounds} 

        \For{embedding $\vec{x}_i$ in newly sampled $\mathcal{X}(\vec{s}_i)$} \label{alg_line:component_assignment_start}
        
            \State Sample $z_i$ for embedding $\vec{x}_i$ using~\eqref{eq:collapsed1}.
            \label{alg_line:fbgmm_inside_loop}
        
        \EndFor \label{alg_line:component_assignment_end}

    \EndFor
\EndFor
\end{algorithmic}
\vspace{-0.6\baselineskip}\hrulefill
\caption{Gibbs sampler for word segmentation and clustering of speech.}\label{alg:gibbs_wordseg}
\end{figure}

More formally, given acoustic data $\{ \vec{s}_i \}_{i = 1}^S$, where every utterance $\vec{s}_i$ consists of acoustic frames $\vec{y}_{1:M_i}$, we need to hypothesize word boundary locations and a word type (mixture component) for each hypothesized segment.
$\mathcal{X}(\vec{s}_i)$ denotes the embedding vectors under the current segmentation for utterance $\vec{s}_i$.
Pseudo-code for the blocked Gibbs sampler, which samples a segmentation utterance-wide, is given in Fig.~\ref{alg:gibbs_wordseg}.
An utterance $\vec{s}_i$ is randomly selected; the embeddings from the current segmentation $\mathcal{X}(\vec{s}_i)$ are removed from the Bayesian GMM; a new segmentation is sampled; and finally the embeddings from this new segmentation are added back into the~Bayesian~GMM.

For each utterance $\vec{s}_i$ a new set of embeddings $\mathcal{X}(\vec{s}_i)$ is sampled in line~\ref{alg_line:sample_bounds} of Fig.~\ref{alg:gibbs_wordseg}.  This is done using the forward filtering backward sampling dynamic programming algorithm~\cite{scott_jasa02}.
Forward variable $\alpha[t]$ is defined as the density of the frame sequence $\vec{y}_{1:t}$, with the last frame the end of a word: $\alpha[t] \defeq p(\vec{y}_{1:t} | {h^-})$.
{The embeddings and component assignments for all words not in $\vec{s}_i$, and the hyperparameters of the GMM, are denoted as $h^- = (\mathcal{X}_{\backslash s}, \vec{z}_{\backslash s}; a, \vec{\beta})$.}
To derive recursive equations for $\alpha[t]$, we use a variable $q_t$ to indicate the number of
{acoustic observation frames in the hypothesized word that ends at}
frame $t$: if $q_t = j$, then $\vec{y}_{t - j + 1:t}$ is a word.
The forward variables can then be recursively calculated as:
\begin{align}
    \alpha[t]
    &= p(\vec{y}_{1:t} | h^-)
    = \sum_{j = 1}^t p(\vec{y}_{1:t}, q_t = j | h^-) \nonumber \\
    &= \sum_{j = 1}^t p(\vec{y}_{{t - j + 1}:t} | h^-) p(\vec{y}_{1:{t-j}}, q_t = j | h^-) \nonumber \\
    &= \sum_{j = 1}^t p(\vec{y}_{{t - j + 1}:t} | h^-) \alpha[t - j] \label{eq:forward}
\end{align}
{starting with $\alpha[0] = 1$ and calculating~\eqref{eq:forward}
for $1 \leq t \leq M - 1$.}

The $p(\vec{y}_{{t - j + 1}:t} | h^-)$ term in~\eqref{eq:forward} is the value of a joint PDF over acoustic frames $\vec{y}_{{t - j + 1}:t}$.
In a frame-based supervised setting, this term would typically be calculated as the product of the PDF values of a GMM (or prior-scaled posteriors of a deep neural network) for the frames involved.
However, we work at a whole-word segment level, and our acoustic model is defined over a whole segment, which means we need to define this term explicitly. 
Let $\vec{x}' = f(\vec{y}_{{t - j + 1}:t})$ be the word embedding calculated on the acoustic frames $\vec{y}_{{t - j + 1}:t}$ (the hypothesized word).
We then treat the term as:
\begin{equation}
    p(\vec{y}_{{t - j + 1}:t} | h^-) \defeq \left[p \left(\vec{x}' | h^- \right) \right]^j \label{eq:prob_segment}
\end{equation}
Thus, as in the frame-based supervised case, each frame is assigned a PDF score.
But instead of having a different PDF value for each frame, all $j$ frames in the segment $\vec{y}_{{t - j + 1}:t}$ are assigned the PDF value of the whole segment under the current acoustic model.
Another interpretation is to see $j$ as a language model scaling factor, used to combine the continuous embedding and discrete unigram spaces.
In initial experiments we found that without this factor, severe over-segmentation occurred.
The marginal term in~\eqref{eq:prob_segment} can be calculated as:
\begin{align}
    p(\vec{x}' | h^-)
    &= \sum_{k = 1}^{K} p(\vec{x}', z_h = k | \mathcal{X}_{\backslash h}, \vec{z}_{\backslash h}; a, \vec{\beta}) \nonumber \\
    &= \sum_{k = 1}^{K} P(z_h = k | \vec{z}_{\backslash h}; a)  p(\vec{x}'| \mathcal{X}_{k\backslash h}; \vec{\beta})
    \label{eq:likelihood_fbgmm}
\end{align}
The two terms in~\eqref{eq:likelihood_fbgmm} are provided by the Bayesian GMM acoustic model, as given in equations~\eqref{eq:first_term7} and~\eqref{eq:second_term1}, respectively.

Once all $\alpha$'s have been calculated, a segmentation can be sampled backwards~\cite{mochihashi+etal_acl09}.
Starting from the final positition $t = M$, we sample the preceding word boundary position~using
\begin{equation}
    P(q_t = j | \vec{y}_{1:t}, h^-) \propto p(\vec{y}_{{t - j + 1}:t} | h^-) \alpha[t - j]
    \label{eq:backward}
\end{equation}
We calculate~\eqref{eq:backward} for $1 \leq j \leq t$ and sample while $t - j \geq 1$.

Fig.~\ref{alg:gibbs_wordseg} gives the complete sampler for our model, showing how segmentation and clustering of speech is performed jointly.
The inner part of Fig.~\ref{alg:gibbs_wordseg} is also illustrated in Fig.~\ref{fig:unsup_wordseg}(a): lines~\ref{alg_line:remove_embeds} to~\ref{alg_line:sample_bounds} perform word segmentation which proceeds from top to bottom in Fig.~\ref{fig:unsup_wordseg}(a), while lines~\ref{alg_line:component_assignment_start} to~\ref{alg_line:component_assignment_end} perform acoustic modelling which proceeds from bottom to top.% in the figure.

\subsection{Iterating the model}
\label{sec:iterating}

As explained in Section~\ref{sec:embeddings}, the fixed-dimensional embedding extraction relies on a reference set $\mathcal{Y}_{\text{ref}}$.
In~\cite{levin+etal_asru13}, this set was composed of true word segments.
In this unsupervised setting, we do not have such a set.
We therefore start with exemplars extracted randomly from the data.
Using this set, we extract embeddings and then run our sampler in an unconstrained setup where it is free to discover an order of magnitude more clusters than the true number of word types.
From the biggest clusters discovered in this first iteration (those that cover 90\% of the data), we extract a new exemplar set, which is used to recalculate embeddings.
We repeat this procedure for a number of iterations, resulting in a refined exemplar set~$\mathcal{Y}_{\text{ref}}$.

\section{Experiments}
\label{sec:experiments}

\subsection{Evaluation setup}
\label{sec:evaluation}

We evaluate using the TIDigits connected digit corpus~\cite{leonard_icassp84}, which has a vocabulary of 11 English digits: `oh' and `zero' through~`nine'.
Using this simple small-vocabulary task, we are able to thoroughly analyze 
the discovered units and report results on the same corpus as 
several previous unsupervised studies~\cite{tenbosch+cranen_interspeech07,sun+vanhamme_csl13,walter+etal_asru13,vanhainen+salvi_icassp14}.
In particular, we use the recent results of Walter et al.~\cite{walter+etal_asru13} as baselines in our own experiments.

TIDigits consists of an official training set with 112~speakers (male and female) and 77 digit sequences per speaker, and a comparable test set.
Each set contains about 3~hours of speech.
Our model is unsupervised, which means that the concepts of training and test data become blurred.
We run our model on both sets separately---in each case, unsupervised modelling and evaluation is performed on the same set.
To avoid confusion with supervised regimes, we relabel the official TIDigits training set as `{TIDigits1}' and the test set as `{TIDigits2}'.
TIDigits1 was used during development for tuning hyperparameters (see Section~\ref{sec:model_implementation}); TIDigits2 was treated as unseen final test set.

For evaluation, the unsupervised decoded output of a system is compared to the ground truth transcriptions. From this comparison a mapping matrix $\vec{G}$ is constructed: 
$G_{ij}$ is the number of acoustic frames that are labelled as digit $i$ in the ground truth transcript and labelled as discovered word type $j$ by the model.
We then use three quantitative evaluation metrics: 
\renewcommand{\IEEEiedlistdecl}{\settowidth{\IEEElabelindent}{}} % {l} would be width of 'l'
\begin{itemize}
    \item \textit{Average cluster purity:} Every discovered word type (cluster) is mapped to the most common ground truth digit in that cluster, given by $i' = \argmax_{i} G_{ij}$ for cluster $j$. Average purity is then defined as the total proportion of the correctly mapped frames: ${\sum_j \max_{i} G_{ij}} / {\sum_{i,j} G_{ij}}$.
If the number of discovered types is more than the true number of types, more than one cluster may be mapped to a single ground truth type (i.e.\ a many-to-one mapping, as in~\cite{sun+vanhamme_csl13}).
    \item \textit{Unsupervised WER:} Discovered types are again mapped, but here at most one cluster is mapped to a ground truth digit~\cite{walter+etal_asru13}. 
By then aligning the mapped decoded output from a system to the ground truth transcripts, we calculate $\textrm{WER} = \frac{S + D + I}{N}$, with $S$ the number of substitutions, $D$ deletions, $I$ insertions, and $N$ the tokens in the ground truth.
In cases where the number of discovered types is greater than the true number, {some clusters
will be left unassigned and counted as errors.}
    \item \textit{Word boundary $F$-score:} By comparing the
word boundary positions proposed by a system to those from
forced alignments of the data (falling within 40 ms), we 
calculate word boundary precision and recall, and report the $F$-scores.
\end{itemize}

We consider two system initialization strategies, which were also used in~\cite{walter+etal_asru13}: (i)~random initialization; and (ii)~initialization from a separate UTD system.
In the UTD condition, the boundary positions and cluster assignments for the words discovered by a UTD system can be used.
Walter et al.\ used both the boundaries and assignments, while we use only the boundaries for initialization 
(we didn't find any gain by using the cluster identities as well).
We use the UTD system of~\cite{jansen+vandurme_asru11}.

As mentioned in Section~\ref{sec:related_studies_full_coverage}, Walter et al.\ constrained their system to only discover 11 clusters (the true number).
For our model we consider two scenarios: (i) in the \textit{constrained} setting, we fix the number of components of the model to $K = 15$; (ii)~in the \textit{unconstrained} setting, we allow the model to discover up to $K = 100$ clusters.
For the first, we use $K = 15$ instead of $11$ since we found that more consistent performance is achieved when allowing some variation in cluster discovery.
In the second setting, $K = 100$ allows the model to discover many more clusters than the true number of types.
Since the Bayesian GMM is able to (and does) empty out some of its components (not all 100 clusters need to be used) this represents the case where we do not know vocabulary size upfront and the model itself is required to find a suitable number of clusters.

\subsection{Model implementation and hyperparameters}
\label{sec:model_implementation}

The hyperparameters of our model are set mainly based on previous work on other tasks~\cite{kamper+etal_slt14}.
However, some parameters were changed by hand during development. 
These changes were made based on performance on {TIDigits1}.
Below, we also note the changes we made from our own preliminary work~\cite{kamper+etal_interspeech15}.
The hyperparameters we used in~\cite{kamper+etal_interspeech15} led to far less consistent performance over multiple sampling runs: WER standard deviations were in the order of 9\% absolute, compared to the deviations of less than 1\% that we obtain in Section~\ref{sec:results}.

For the acoustic model (Section~\ref{sec:gmm}), we use the following hyperparameters, based on~\cite{kamper+etal_slt14,murphy_bayesgauss07,wood+black_jnm08}: all-zero vector for $\vec{\mu}_0$, $a = 1$, $\sigma^2 = 0.005$, $\sigma_0^2 = \sigma^2/\kappa_0$ and $\kappa_0 = 0.05$.
Based on~\cite{levin+etal_asru13,kamper+etal_slt14} we use the following parameters for the fixed-dimensional embedding extraction (Section~\ref{sec:embeddings}): dimensionality $D = 11$, $k = 30$, $\sigma_K = 0.04$, $\xi = 2.0$ and $N_\text{ref}=8000$.
The embedding dimensionality for this small-vocabulary task is less than that typically used for other larger-vocabulary unsupervised tasks (e.g.\ $D = 50$ in \cite{kamper+etal_slt14}).
In our preliminary work on TIDigits~\cite{kamper+etal_interspeech15}, we used $D = 15$ with $N_\text{ref}=5000$, but here we found that using $D = 11$ with a bigger reference set $N_\text{ref}=8000$ gave more consistent performance on TIDigits1.
For embedding extraction, speech is parameterized as 15-dimensional frequency-domain linear prediction features~\cite{athineos+ellis_asru03} at a frame rate of 10~ms, and cosine distance is used as similarity metric in DTW alignments.

As in~\cite{kamper+etal_slt14}, embeddings are normalized to the unit sphere.
We found that some embeddings were close to zero, causing issues in the sampler.
We therefore add low-variance zero-mean Gaussian noise before normalizing: the standard deviation of the noise is set to $0.05 \cdot \sigma_E$, where $\sigma_E$ is the sample standard deviation of all possible embeddings. 
Changing the $0.05$ factor within the range $[0.01, 0.1]$ made little difference.

In Section~\ref{sec:iterating} we explained that to find the reference set $\mathcal{Y}_\textrm{ref}$ for embedding extraction, we start with exemplars extracted randomly from the data, and then iteratively refine the set by using the decoded output from our model.
In the first iteration we use $N_\text{ref}=8000$ random exemplars.
In subsequent iterations, we use terms from the biggest discovered clusters that cover at least 90\% of the data: 
we use the word tokens with the highest marginal densities as given by~\eqref{eq:likelihood_fbgmm} in each of these clusters to yield $4000$ discovered exemplars which we use in addition to $4000$ exemplars again extracted randomly from the data, to give a total set of size $N_\text{ref}=8000$.
We found that performance was more consistent when still using some random exemplars in $\mathcal{Y}_\textrm{ref}$ after the first iteration.

To make the search problem in Fig.~\ref{alg:gibbs_wordseg} tractable, we require potential words to be between 200~ms and 1~s in duration, and we only consider possible word boundaries at 20~ms intervals.
By doing this, the number of possible embeddings is greatly reduced.
{Although embedding comparisons are fast, the calculation of the embeddings is not, and this is the main bottleneck of our approach.}
In our implementation, all allowed embeddings are pre-computed.
The sampler can then look up a particular embedding without the need to compute it on the~fly.

% /endgame/python/projects/phd/segmentalist/unigram_segmentalist_exp_tidigits/doc/plots/plot_mappings.py
\begin{figure*}[!t]
    \centering
    {\footnotesize (a)}~\includegraphics[scale=0.58]{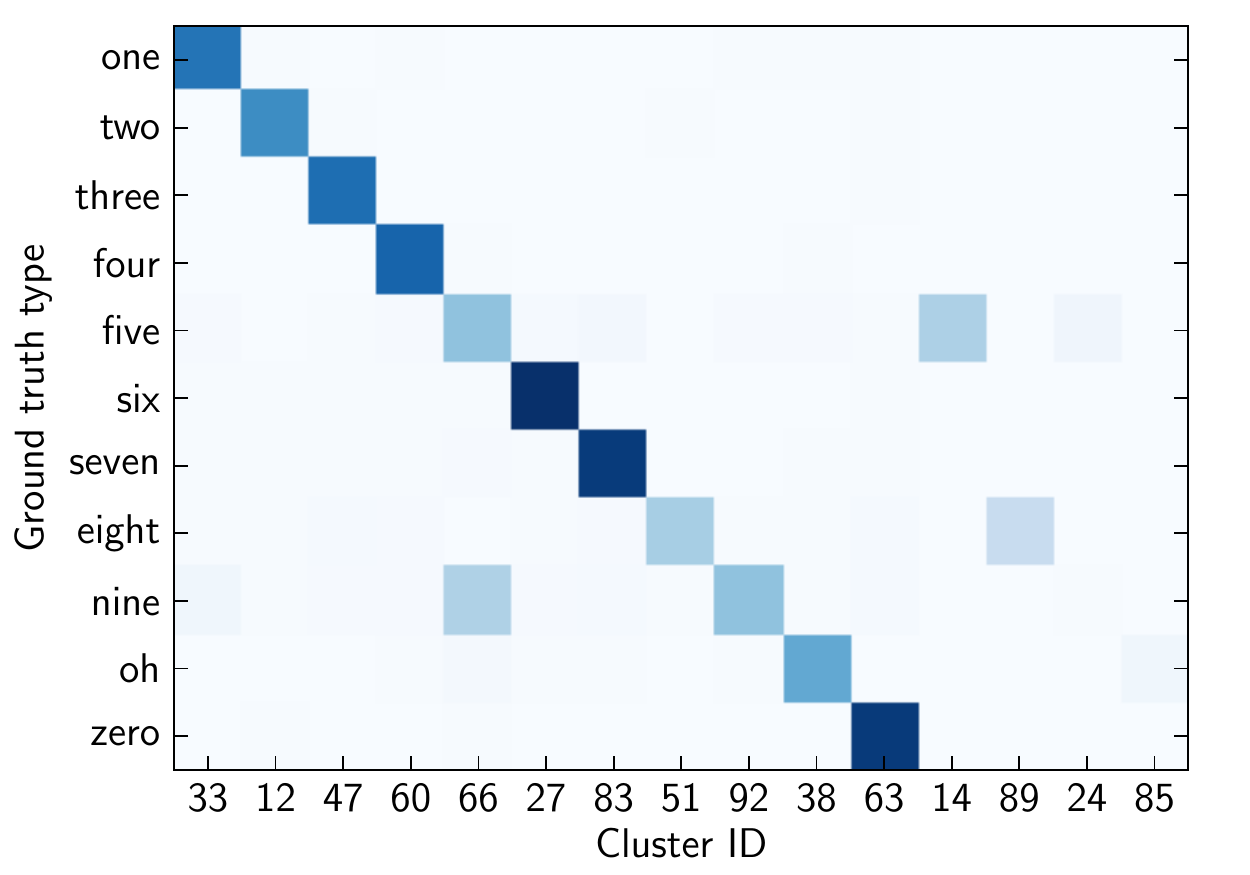} \hspace{2em}
    {\footnotesize (b)}~\includegraphics[scale=0.58]{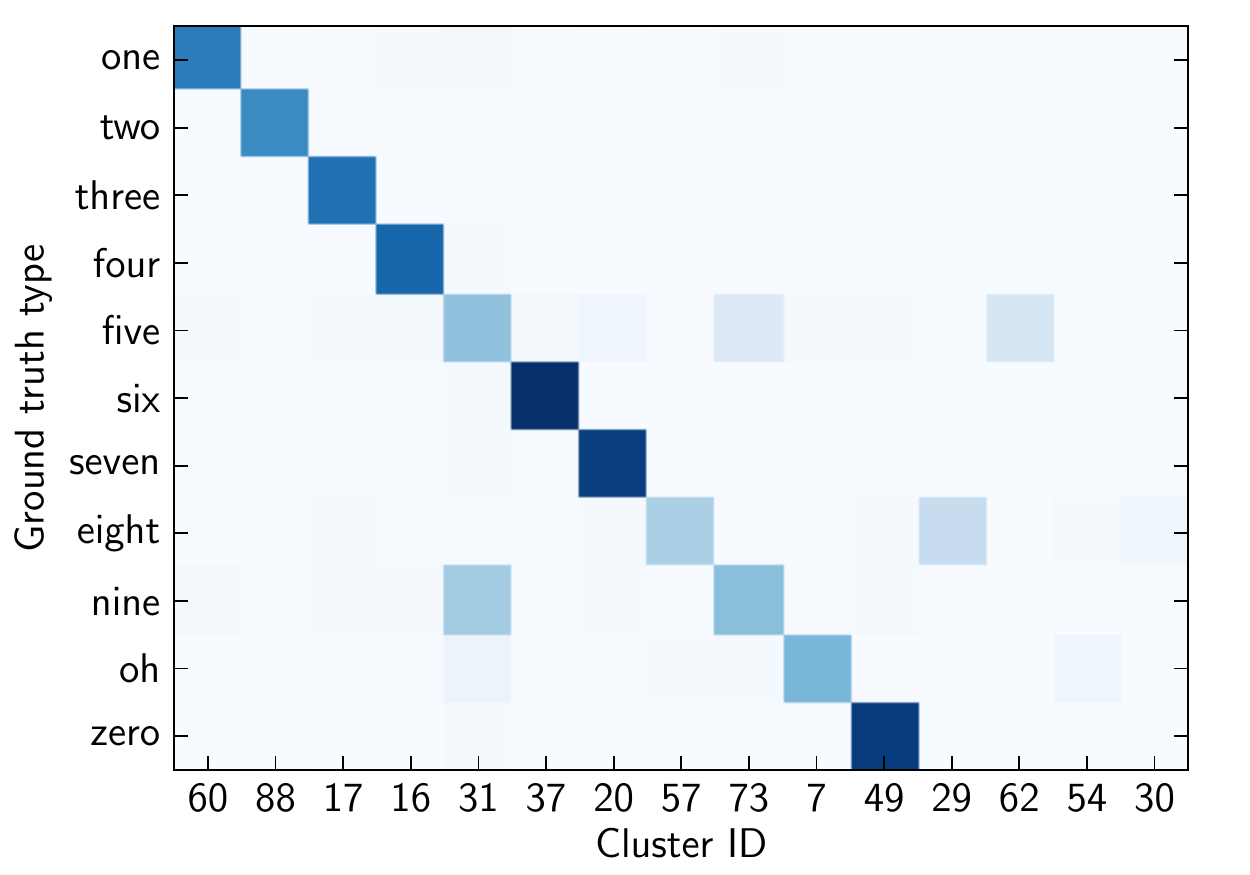}
   
    \caption{Mapping matrices between ground truth digits and discovered word types for (a) the third and (b) the fifth iteration unconstrained models in Table~\ref{tbl:exemplars}.% and (b) the fifth-iteration model.
    }
    \label{fig:unconstrained_mappings}
\end{figure*}

% /media/kamperh/endgame/python/projects/phd/segmentalist/unigram_segmentalist_exp_tidigits/doc/results.ods
\begin{table}[!b]
    \mytable
    \caption{Performance of the Unconstrained Segmental Bayesian Model on TIDigits1 as the Reference Set is Refined%.
    }
    \begin{tabularx}{\linewidth}{@{}c||C|C|C|C|C@{}}
        \hline
        Iteration                       & 1 & 2 & 3 & 4 & 5 \\
        \hline \hline
        WER (\%)                        & $35.4$ & $23.5$ & $21.5$ & $21.2$ & $22.9$ \\
        Avg.\ cluster purity (\%)       & $86.5$ & $89.7$ & $89.2$ & $88.5$ & $86.6$ \\
        Bound.\ $F$-score (\%)         & $70.6$ & $72.2$ & $71.8$ & $70.9$ & $69.4$ \\
        Clusters covering 90\%              & 20             & 13 & 13 & 13 & 13 \\
        \hline
    \end{tabularx}
    \label{tbl:exemplars}
\end{table}

To improve sampler convergence, we use simulated annealing~\cite{goldwater+etal_cognition09}, by raising the boundary probability in~\eqref{eq:backward} to the power $\frac{1}{\gamma}$ before sampling, where $\gamma$ is a temperature parameter.
We also found that convergence is improved by first running the sampler in Fig.~\ref{alg:gibbs_wordseg} without sampling boundaries.
In all experiments we do this for 25 iterations.
Subsequently, the complete sampler is run for $J = 25$ Gibbs sampling iterations with 5 annealing steps in which $\frac{1}{\gamma}$ is increased linearly from $0.01$ to $1$.
In all cases we run 5 sampling chains in parallel~\cite{resnik+hardisty_gibbs_tutorial10}, and report average performance and standard deviations.

\subsection{Results and analysis}
\label{sec:results}

\textit{Unconstrained model evaluation:}

As explained, we use our model to iteratively rediscover the embedding reference set $\mathcal{Y}_{\text{ref}}$.
Table~\ref{tbl:exemplars} shows the performance of the unconstrained segmental Bayesian model on TIDigits1 as the reference set is refined.
Random initialization is used throughout.
Unconstrained modelling represents the most realistic setting where vocabulary size is not known upfront.
Standard deviations were less than $0.3\%$ absolute for all~metrics.

Despite being allowed to discover many more clusters (up to 100) than the true number of word types (11), the model achieves a WER of 35.4\% in the first iteration, which improves to around 21\% in iterations 3 and 4.
Error rate increases slightly in iteration 5.
Cluster purity over all iterations is above $86.5\%$, which is higher than the scores of around 85\% reported by Sun and Van hamme~\cite{sun+vanhamme_csl13}. 
Word boundary $F$-scores are around 70\% over all iterations.
As mentioned, the Bayesian GMM is biased not to use all of its 100 components.
Despite this, none of the models empty out any of their components.
However, most of the data is covered by only a few components: the last row in Table~\ref{tbl:exemplars} shows that in the first iteration, 90\% of the data is covered by the 20 biggest mixture components, while this number drops to 13 clusters in subsequent iterations.

In order to analyze the type of errors that are made, we visualize the mapping matrix $\vec{G}$, which gives the number of frames of overlap between the ground truth digits and the discovered word types (Section~\ref{sec:evaluation}).
Fig.~\ref{fig:unconstrained_mappings} shows the mappings for the 15 biggest clusters 
of the unconstrained models of iterations 3 and 5 of Table~\ref{tbl:exemplars}, respectively.

Consider the mapping in Fig.~\ref{fig:unconstrained_mappings}(a) for iteration 3.
Qualitatively we observe a clear correspondence between the ground truth and discovered word types, which coincides with the high average purity of 89.2\%.
Apart from cluster 66, all other clusters overlap mainly with a single digit.
Listening to cluster 66 reveals that most tokens correspond to [ay v] from the end of the digit `five' and tokens of [ay n] from the end of the `nine', both dominated by the diphthong. 
Correspondingly, most of the tokens in cluster 14 are the beginning [f ay] of `five', while cluster 92 is mainly the beginning [n~ay] of `nine'.
The digit `eight' is split across two clusters: cluster 51 mainly contains `eight' tokens where the final [t] is not pronounced, 
while in cluster 89 the final [t] is explicitly produced.

Table~\ref{tbl:exemplars} shows that performance deteriorates slightly in iteration 5. % of reference set refinement.
{By comparing Figures~\ref{fig:unconstrained_mappings}(a) and (b), the source of the extra errors can be observed:}
overall the mapping in the fifth iteration (b) looks similar to that of the third (a), except the digit `five' is now also partly covered  by a third cluster (73).
This cluster mainly contains beginning portions of `five' and `nine', again dominated by the diphthong [ay].
Cluster 62 in this case mainly contains tokens of the fricative [f] from `five'.
Note that both WER and boundary $F$-score penalize the splitting of digits, although the discovered clusters correspond to consistent partial words.
Below we discuss this issue further.

% /endgame/python/projects/phd/segmentalist/unigram_segmentalist_exp_tidigits/doc/plots/plot_discovered_embeds.py
\begin{figure}[!b]
    \centering
    \includegraphics[scale=0.58]{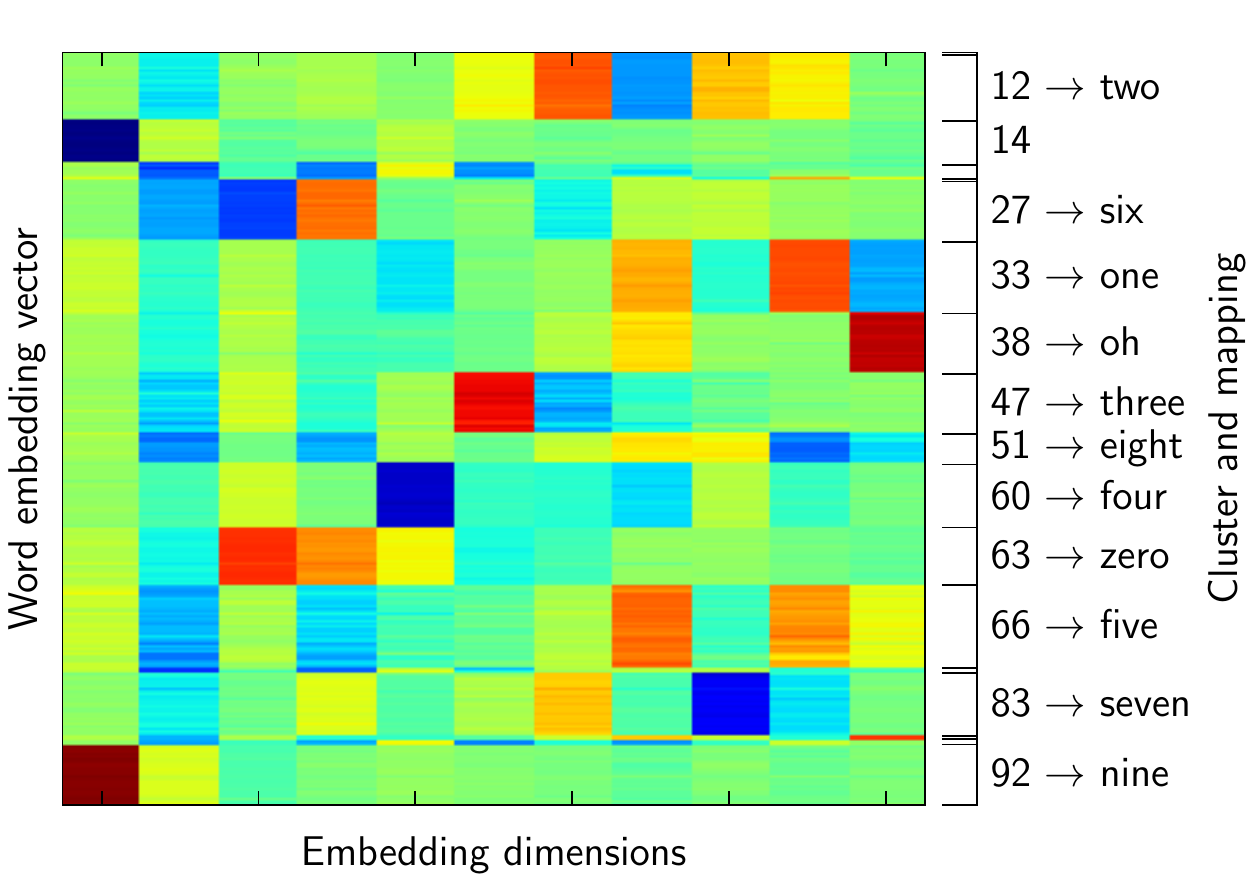}
    \caption{Embedding vectors for the discovered word types from a single speaker for the iteration 3 unconstrained model in Table~\ref{tbl:exemplars}. The greedy mapping from discovered to true word type (for calculating WER) is given on the right; since there are more clusters than digits, some clusters are left unmapped.}
    \label{fig:discovered_embeddings}
\end{figure}

One might suspect from the analysis in Fig.~\ref{fig:unconstrained_mappings} that some of the
discovered word types are bi-modal, i.e. that when a single component
of the Bayesian GMM contains two different true types  (e.g. cluster
66 in Fig.~\ref{fig:unconstrained_mappings}(a), with tokens of both `five' and `nine'), there might
be two relatively distinct sub-clusters of embeddings within that
component. However, this is not the case. Fig.~\ref{fig:discovered_embeddings} shows the embeddings
of the discovered word types for a single speaker from the model in
iteration 3 of Table~\ref{tbl:exemplars}; embeddings are ordered and stacked by
discovered type along the $y$-axis, with the embedding values coloured
along the $x$-axis.  The embeddings for cluster 66 appear uni-modal,
despite containing both [ay v] and [ay n] tokens; yet they are
distinct from the embeddings in cluster 92 ([n ay] tokens) and cluster
14 ([f ay]). This analysis suggests that the model is finding sensible
clusters given the embedding representation it has, and to
consistently improve results we would need to focus on developing more
discriminative embeddings. %---an interesting area for future work.

% /media/kamperh/endgame/python/projects/phd/segmentalist/unigram_segmentalist_exp_tidigits/doc/results.ods
\begin{table}[!b]
    \mytable
    \caption{WER (\%) on TIDigits1 of the Unsupervised Discrete HMM System of Walter et al.~\cite{walter+etal_asru13} and the Segmental Bayesian Model
    }
    \begin{tabularx}{\linewidth}{c|c||C|C}
        \hline
        Model & Constrained  & Random init. & UTD init. \\ %& Oracle \\
        \hline \hline
        Discrete HMM~\cite{walter+etal_asru13} & yes & 32.1 & 18.1 \\ % & \textbf{11.9} \\
        Segmental Bayesian & yes & $19.4 \pm 0.3$ & $19.4 \pm 0.1$ \\ % & 12.1 \\
        Segmental Bayesian & no & $21.5 \pm 0.1$ & -  \\ % &  -  \\
%         Discrete HMM~\cite{walter+etal_asru13} & \checkmark & 67.9 & 81.9 & \textbf{88.1} \\
%         Segmental Bayesian & \checkmark & \textbf{88.8} & \textbf{87.9} & 87.9 \\
%         Segmental Bayesian & $\times$ & 79.4 & -  & -  \\
        \hline
    \end{tabularx}
    \label{tbl:train_results}
\end{table}

\begin{table*}[!t]
    \mytable
    \caption{Performance of the Bayesian Segmental Model on TIDigits1 and TIDigits2, with Random Initialization%, the Latter Completely Unseen
    }
    \begin{tabularx}{\linewidth}{@{}c||C|C|c|C|C|c@{}}
        \hline
        \multirow{2}{*}{Model} & \multicolumn{3}{c|}{TIDigits1 (\%)} & \multicolumn{3}{c}{TIDigits2 (\%)} \\
        \cline{2-7}
        & WER & Cluster purity & Boundary $F$-score & WER & Cluster purity & Boundary $F$-score \\
        \hline \hline
%         Discrete HMM~\cite{walter+etal_asru13} & 32.1 & - & - & - &  - & - \\
%         \hline
        Constrained segmental Bayesian & $19.4 \pm 0.3$ & $88.4 \pm 0.06$ & $70.6 \pm 0.2$ & $13.2 \pm 1.0$ & $91.2 \pm 0.2$ & $76.7 \pm 0.7$ \\
        Unconstrained segmental Bayesian & $21.5 \pm 0.1$ & $89.2 \pm 0.1$ & $71.8 \pm 0.2$ & $17.6 \pm 0.2$ & $92.5 \pm 0.1$ & $77.6 \pm 0.3$\\
        \hline
    \end{tabularx}
    \label{tbl:train_test_results}
\end{table*}

\textit{Constrained model evaluation and comparison:}

To compare with the discrete HMM-based system of Walter et al.~\cite{walter+etal_asru13}, we use the exemplar set discovered in  
iteration 3 of Table~\ref{tbl:exemplars} (using an unconstrained setup up to this point) and then constrain the Bayesian segmental model to 15 components.
Table~\ref{tbl:train_results} shows WERs achieved on TIDigits1. 
Under random initialization, the constrained segmental Bayesian model performs $12.7\%$ absolute better than the discrete HMM.
When using UTD for initialization, the discrete HMM does better by $1.3\%$ absolute.
The WER of the third-iteration unconstrained model in Table~\ref{tbl:exemplars} is repeated in the last row of Table~\ref{tbl:train_results}.
Despite only mapping 11 out of 100 clusters to true labels, this unconstrained model still yields 10.6\% absolute lower WER  than the randomly-initialized discrete HMM with the correct number of clusters.
By comparing rows 2 and 3, we observe that there is only a $2.1\%$ absolute gain in WER by constraining the Bayesian model to a stricter number of types.

\textit{Generalization and hyperparameters:}

As noted in Section~\ref{sec:model_implementation}, some development decisions were made based on performance on TIDigits1.
TIDigits2 was kept as unseen data up to this point.
Using the setup developed on TIDigits1, we repeated exemplar extraction and segmentation separately on TIDigits2.
Three iterations of exemplar refinement were used.
Table~\ref{tbl:train_test_results} shows the performance of randomly-initialized systems on both TIDigits1 and TIDigits2, with the performance on TIDigits1 repeated from Table~\ref{tbl:train_results}.

\begin{figure}[!t]
    \centering
    \includegraphics[scale=0.58]{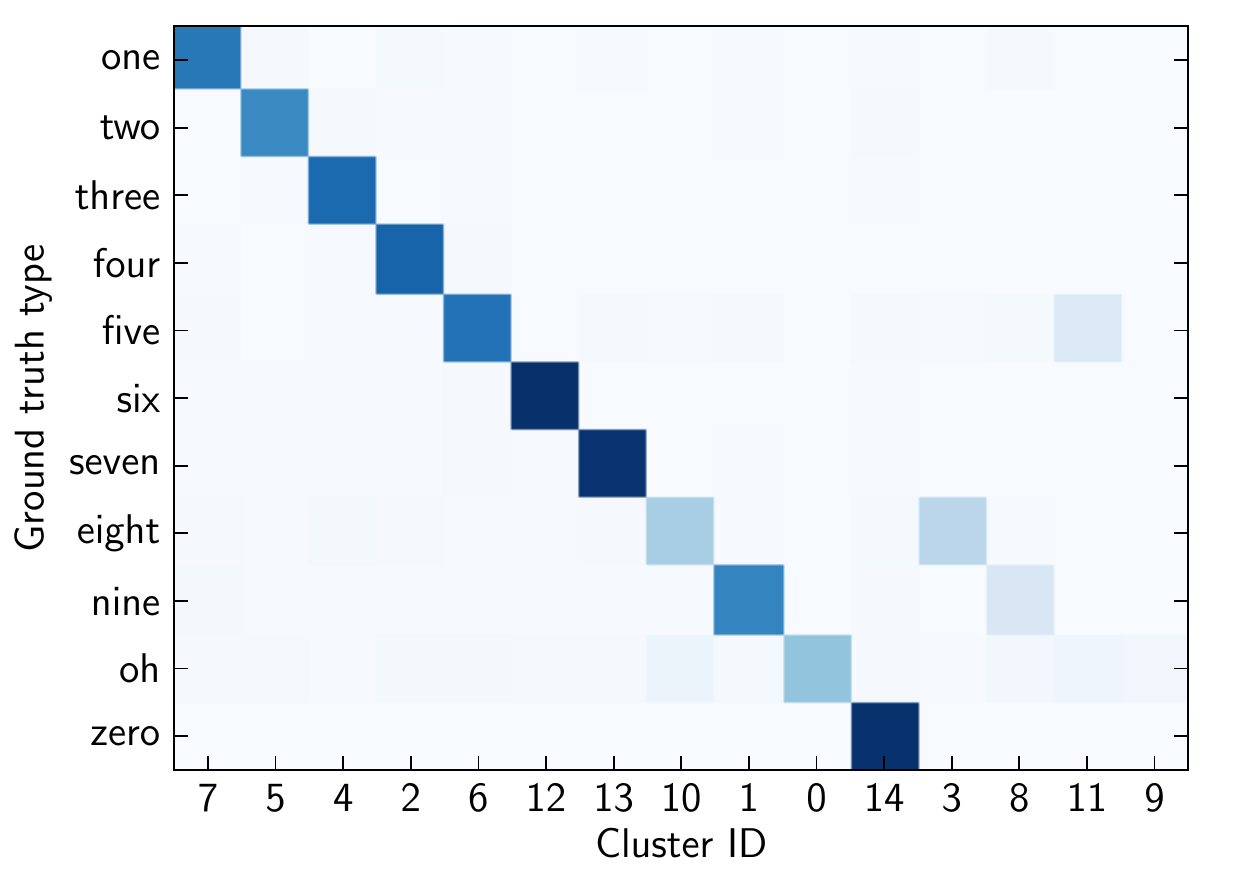}
    \caption{Mapping matrix between ground truth digits and discovered word types for the constrained segmental Bayesian model in Table~\ref{tbl:train_test_results} on TIDigits2.}
    \label{fig:mapping_test_v8_4_pass_3_am_K_15}
\end{figure}

Across all metrics, performance is better on TIDigits2 than on TIDigits1: WERs drop by 6.2\% and 3.9\% absolute for the constrained and unconstrained models, respectively; cluster purity improves by around 3\% absolute; and boundary $F$-score is higher by 6\% absolute.
To understand this discrepancy, consider the mapping matrix in Fig.~\ref{fig:mapping_test_v8_4_pass_3_am_K_15} for the constrained segmental Bayesian model on TIDigits2 (13.2\% WER, Table~\ref{tbl:train_test_results}).
The figure shows that every cluster is dominated by data from a single ground truth digit. Furthermore, all digits apart from `eight' are found in a single cluster.
Now consider the mapping in Fig.~\ref{fig:unconstrained_mappings}(a) for the unconstrained segmental Bayesian model on TIDigits1 (giving the higher WER of 21.5\%, Table~\ref{tbl:train_test_results}).
This mapping is similar to that of Fig.~\ref{fig:mapping_test_v8_4_pass_3_am_K_15}, apart from two digits: both `five' and `nine' are split into two clusters, corresponding to beginning and end partial words.
Although these digits are consistently decoded as the same sequence of clusters, WER counts the extra clusters as insertion errors.
These small differences in the discovered word types results in a non-negligible difference in WER between TIDigits1 and TIDigits2.

% endgame/python/projects/phd/segmentalist/unigram_segmentalist_exp_tidigits/doc/plots/plot_mappings.py

This analysis and our previous discussion of Fig.~\ref{fig:unconstrained_mappings} indicate that 
unsupervised WER is a particularly harsh measure of unsupervised word segmentation performance: the model may discover consistent units, but if these units do not coincide with whole words, the system will be penalized.
This is also the case for word boundary $F$-score.
Average cluster purity is less affected since a many-to-one mapping is performed; Table~\ref{tbl:train_test_results} shows that purity changes the least of the three metrics when moving from TIDigits1 to TIDigits2.

\begin{figure}[!t]
    \centering
    \includegraphics[scale=0.58]{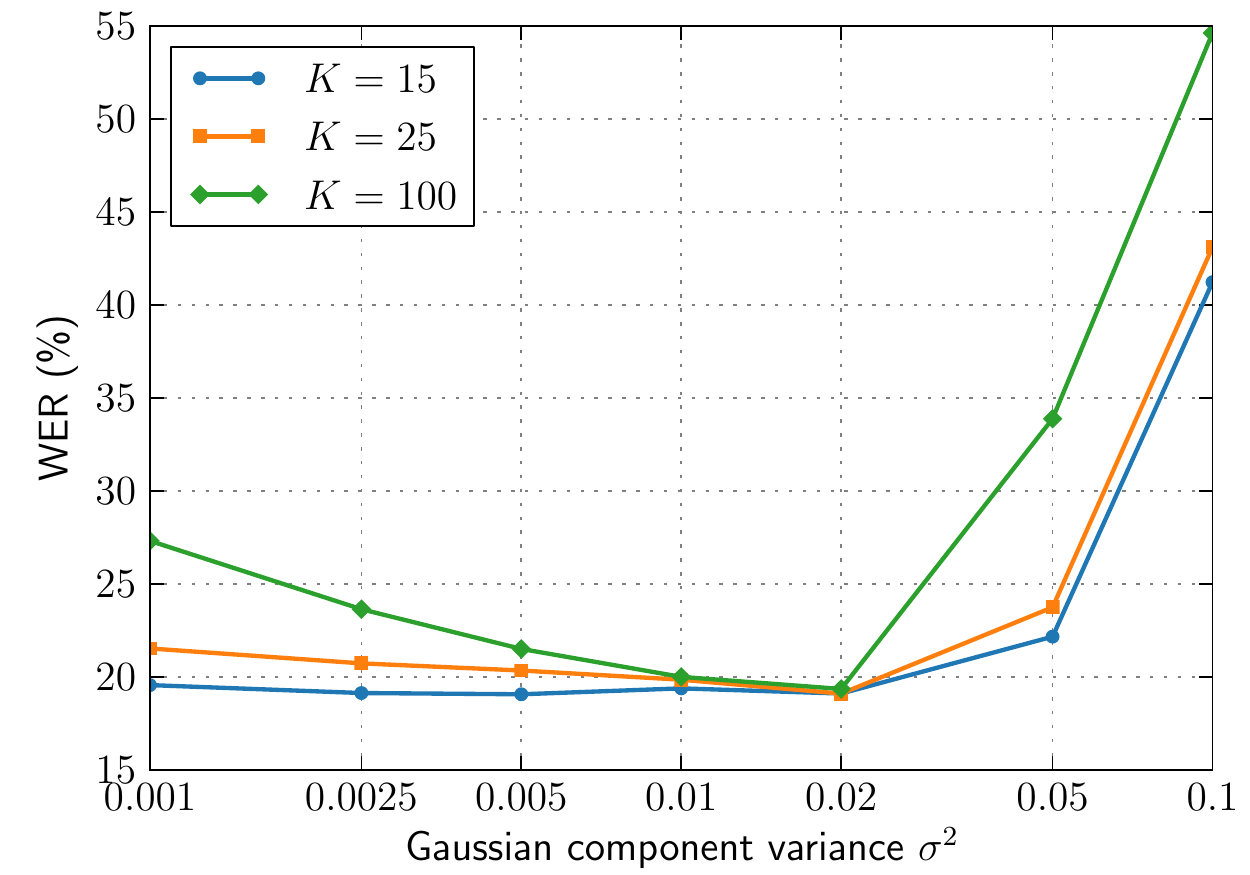}
    \caption{WERs of the segmental Bayesian model on TIDigits1 as the number of Gaussian components $K$ and variance $\sigma^2$ is varied (log-scale on $x$-axis).}
    \label{fig:train_optimize_wer}
\end{figure}

In a final set of experiments, we considered the effect of model hyperparameters.
We found that performance is most sensitive to changes in the maximum number of allowed Gaussian components $K$ and the component variance $\sigma^2$.
Fig.~\ref{fig:train_optimize_wer} shows the effect on WER when changing these hyperparameters.
Results are reasonably stable for $\sigma^2$ in the range $[0.0025, 0.02]$, with WERs below 25\%.
When allowing many components ($K = 100$) and using a small variance, as on the left of the figure, fragmentation takes place with digits being separated into several clusters.
On the right side of the figure, where large variances are used, a few garbage clusters start to capture the majority of the data, leading to poor performance.
The figure also shows that lower WER could be achieved by using a $\sigma^2 = 0.02$ instead of $0.005$ (which we used in the experiments above, based on~\cite{kamper+etal_slt14}).
The reason for the three curves meeting at this $\sigma^2$ setting is that, for all three settings of $K$, more than 90\% of the data are captured by only the 11 biggest clusters.

% /endgame/python/projects/phd/segmentalist/unigram_segmentalist_exp_tidigits/doc/plots/plot_train_optimize.py

\begin{figure}[!t]
    \centering
    \includegraphics[scale=0.58]{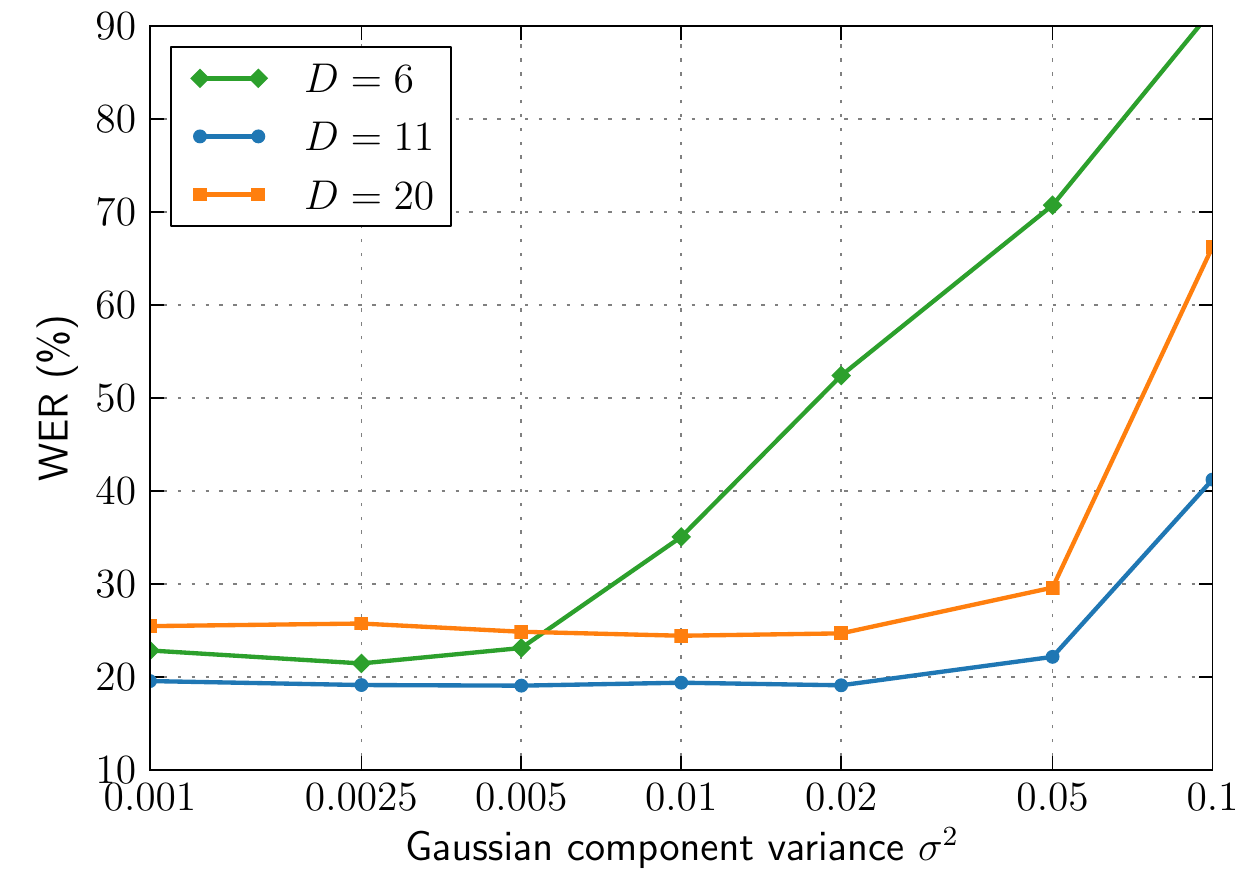}
    \caption{WERs of the segmental Bayesian model on TIDigits1 as the embedding dimensionality $D$ and variance $\sigma^2$ is varied (log-scale on $x$-axis; $K = 15$).}
    \label{fig:train_optimize_dims_wer}
\end{figure}

{We similarly varied the target embedding dimensionality $D$ using a constrained setup ($K = 15$), as shown in Fig.~\ref{fig:train_optimize_dims_wer}. %; the performance of three constrained systems ($K = 15$) is shown in Fig.~\ref{fig:train_optimize_dims_wer}.
For $D = 6$, garbage clusters start to capture the majority of the tokens at lower settings of $\sigma^2$ than for $D = 11$ and $D = 20$.
Much more stable performance is achieved in the latter two cases.
The slightly worse performance of the $D = 20$ setting compared to the others is mainly due to a cluster containing the diphthong [ay], which is present in both `five' and `nine'.%, again highlighting the harshness of WER as unsupervised evaluation metric.
}

% /endgame/python/projects/phd/segmentalist/unigram_segmentalist_exp_tidigits/doc/plots/plot_train_optimize_dims.py

\section{Challenges in Scaling to Larger Vocabularies}
\label{sec:scaling}

We evaluated our system on a small-vocabulary dataset in order to compare to previous work and to allow us to thoroughly analyze the discovered structures.
Our long-term aim (shared by many of the researchers mentioned in Section~\ref{sec:related_studies_full_coverage}) is to scale our system to more realistic multi-speaker corpora with larger vocabularies. % on a larger multi-speaker corpus.
Here we discuss the challenges in doing~so.

\begin{figure}[!t]
    \centering
    \includegraphics[scale=0.5]{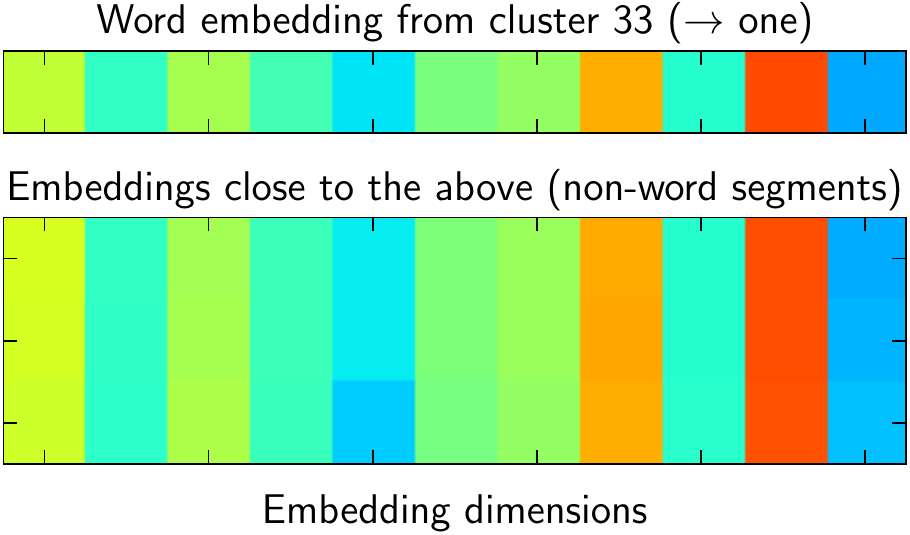}
    \caption{Four embeddings from the same speaker as in Fig.~\ref{fig:discovered_embeddings}: the top one is from cluster 33, the bottom three are from short non-word speech segments.}% not overlapping with `one'.}
    \label{fig:nearest_neighbour_one}
\end{figure}

The fixed-dimensional embedding calculations are the main bottleneck in our overall approach, since embeddings must be computed for each of the very large number of
potential word segments. % considered by the sampler (Fig.~\ref{alg:gibbs_wordseg}).
The embeddings also limit accuracy; one case in particular where the embedding function produces poor embeddings is for  very short speech segments.
An example is given in Fig.~\ref{fig:nearest_neighbour_one}.
The first embedding is from cluster 33 in Fig.~\ref{fig:discovered_embeddings}, which is reliably mapped to the digit `one'.
The bottom three embeddings are from short segments not overlapping with the true digit `one', with
% : two are almost silence, the other a partial [f].
respective durations 20~ms, 40~ms and 80~ms.
Although these three speech segments have little similarity to the segments in cluster 33, Fig.~\ref{fig:nearest_neighbour_one} shows that their embeddings are a good fit to this cluster.
This is possibly due to the aggressive warping in the DTW alignment of these short segments, together with artefacts from normalizing the embeddings to the unit sphere. % (which, in affect, scales the small DTW values).
This failure-mode is easily dealt with by setting a minimum duration constraint (Section~\ref{sec:model_implementation}), but again shows our model's reliance on accurate embeddings.
% . %, without which over-segmentation occurs.
% Nevertheless, this again illustrates our model's reliance on accurate embeddings.%: if the embedding function has any failure-mode where non-word segments are consistently mapped to similar embeddings, the model will propose these as word~types.
% }

To scale to larger corpora, both the efficiency and accuracy of the embeddings would therefore need to be improved (see~\cite{kamper+etal_arxiv15} for recent supervised efforts in this direction).
More importantly, the above discussion highlights a shortcoming of our approach: the sampler considers potential word boundaries at any position, % (within the duration constraints),
without regard to the original acoustics or any notion of a minimal unit.
Many of the previous studies~\cite{walter+etal_asru13,lee_phd14,lee+etal_tacl15,rasanen+etal_interspeech15} use a first-pass method to find positions of high acoustic change and then only allow word boundaries at these positions.
This implicitly defines a minimal unit: the pseudo-phones or pseudo-syllables segmented in the first pass.
% R{\"a}s{\"a}nen et al.~\cite{rasanen+etal_interspeech15} recently used a similar approach by first discovering pseudo-syllable units.
By using such a first-pass method in our system, % (e.g.\ finding pseudo-syllable boundaries),
the number of embedding calculations would greatly be reduced and it would provide a more principled way to deal with artefacts from short segments. % (since it would define the smallest modelling unit).

Another challenge when dealing with larger vocabularies is the choice of the number of clusters $K$.
% In our experiments, the unconstrained model (Section~\ref{sec:evaluation}) was free to discover an order of a magnitude more clusters (100) than the true number of types (11).
An upper-bound of $K = 100$, as we use for our unconstrained model, would not be sufficient for realistic vocabularies.
% On a large vocabulary an upper-bound of $K = 100$, as we use for the unconstrained model, would not be sufficient.
However, the Bayesian framework would allow us to make our model non-parametric: the Bayesian GMM could be replaced by an infinite GMM~\cite{rasmussen_nips99} which infers the number of clusters automatically.
% We presented initial experiments in this direction in~\cite{kamper+etal_slt14}.

% python/projects/phd/segmentalist/unigram_segmentalist_exp_tidigits/doc/plots/nearest_neighbour/nearest_neighbour_one.py

Finally, in this study we made a unigram word predictability assumption (Section~\ref{sec:word_segmentation}) since the digit sequences do not have
any word-word dependencies. However, in a realistic corpus, such
dependencies will exist and could prove useful (even essential) for
segmentation and lexicon discovery. In particular,~\cite{elsner+etal_emnlp13} showed
that for joint segmentation and clustering of noisy phone sequences, %unigram and bigram models had similar segmentation
%performance, but 
a bigram model was needed to improve
clustering accuracy. %  A bigram model could be even more critical for
% clustering word embeddings.}%, with many potentially overlapping
% clusters.}
% \herman{
Following~\cite{goldwater+etal_cognition09,mochihashi+etal_acl09} it is mathematically straightforward to extend the
algorithm of Section~\ref{sec:word_segmentation} to more complex language models. Exact
computation of the extended model will be slow (e.g. the bigram
extension of equation~\eqref{eq:likelihood_fbgmm} requires marginalizing over the
cluster assignment of both the current and preceding embeddings) but we anticipate that reasonable
approximations will be possible (e.g. only marginalizing over a
handful of the most probable clusters). The development of these extensions and
% extended model and the necessary
approximations is an important part
of our future work on larger vocabularies.

\section{Conclusion}

We introduced a novel Bayesian model, operating on fixed-dimensional embeddings of speech, which segments and clusters unlabelled continuous speech into hypothesized word units---an approach which is very different from any presented before.
We applied our model to a small-vocabulary digit recognition task and compared performance to a more traditional HMM-based approach of a previous study.
Our model outperformed the baseline by more than 10\% absolute in unsupervised word error rate (WER), without being constrained to a small number of word types (as the HMM was).
Analysis showed that our model is reliant on the whole-word fixed-dimensional segment representation: when partial words are consistently mapped to a similar region in embedding space, the model proposes these as separate word types.
Most of the errors of the model were therefore due to consistent splitting of particular digits into partial-word clusters, or separate clusters for the same digit based on pronunciation variation.
The model, however, is not restricted to a particular embedding method.
Future work will investigate more accurate and efficient embedding approaches {and unsupervised language modelling}.

\bibliographystyle{IEEEtran}
\bibliography{taslp2015}

\begin{IEEEbiography}[{\includegraphics[width=1in,height=1.25in,clip,keepaspectratio]{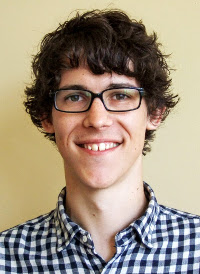}}]{Herman Kamper}
received the B. (Eng.) and M.Sc. (Eng.) degrees in electrical and electronic engineering from Stellenbosch University, Matieland, South Africa in 2009 and 2012, respectively. From 2012-2013 he was research associate and then lecturer of applied mathematics at Stellenbosch University. He is currently pursuing his Ph.D. degree in Informatics within the Centre for Speech Technology Research (CSTR) at the University of Edinburgh, UK.
His main interests are in machine learning and language processing, with a particular interest in the application of unsupervised methods to speech processing problems.
\end{IEEEbiography}

\begin{IEEEbiography}[{\includegraphics[scale=10,width=1in,height=1.25in,clip,keepaspectratio]{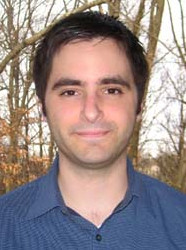}}]{Aren Jansen} is a Senior Research Scientist in the Machine Hearing Group at Google. Before that he was a Senior Research Scientist in the Human Language Technology Center of Excellence and an Assistant Research Professor in the Center for Language and Speech Processing, both at Johns Hopkins University. Aren received the B.A. degree in Physics from Cornell University in 2001. He received the M.S. degree in Physics as well as the M.S. and Ph.D. in Computer Science from the University of Chicago in 2003, 2005, and 2008, respectively. His research has explored a wide range of speech and audio processing topics involving unsupervised/semi-supervised machine learning and scalable retrieval algorithms.
\end{IEEEbiography}

\begin{IEEEbiography}[{\includegraphics[scale=10,width=1in,height=1.25in,clip,keepaspectratio]{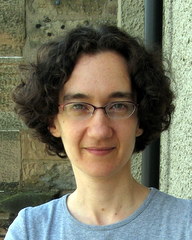}}]{Sharon Goldwater} received an Sc.M. in Computer Science and Ph.D. in
Linguistics from Brown University, Providence, RI, in 2005 and
2007, respectively. She worked as a researcher in the Artificial
Intelligence Laboratory at SRI International from 1998-2000, and
a postdoctoral researcher at Stanford University from
2006-2008. She has worked in the School of Informatics at the
University of Edinburgh since 2008, where she is currently a
Reader in the Institute for Language, Cognition, and Computation (ILCC).
Her research focuses on unsupervised learning for automatic speech and
language processing, and on computer modeling of language acquisition
in children. She has worked most extensively on Bayesian
approaches to the induction of linguistic structure, ranging from
phonetic categories to morphology, syntax, and semantics.
\end{IEEEbiography}

\end{document}